# Anomaly Detection Using One-Class SVM for Logs of Juniper Router Devices

Tat-Bao-Thien Nguyen[1], Teh-Lu Liao[2] and Tuan-Anh Vu[1]

[1] Posts and Telecommunications Institute of Technology, Ho Chi Minh City, Vietnam
`nguyentatbaothien@gmail.com`
[2] Department of Engineering Science, National Cheng Kung University, Tainan, Taiwan

**Abstract.** The article deals with anomaly detection of Juniper router logs. Abnormal Juniper router logs include logs that are usually different from the normal operation, and they often reflect the abnormal operation of router devices. To prevent router devices from being damaged and help administrator to grasp the situation of error quickly, detecting abnormal operation soon is very important. In this work, we present a new way to get important features from log data of Juniper router devices and use machine learning method (basing on One-Class SVM model) for anomaly detection. One-Class SVM model requires some knowledge and comprehension about logs of Juniper router devices so that it can analyze, interpret, and test the knowledge acquired. We collect log data from a lot of real Juniper router devices and classify them based on our knowledge. Before these logs are used for training and testing the One-Class SVM model, the feature extraction phase for these data was carried out. Finally, with the proposed method, the system errors of the routers were dectected quickly and accurately. This may help our company to reduce the operation cost for the router systems.

**Keywords:** Anomaly Detection, Juniper Devices, One-Class SVM, Log Feature Extraction.

## 1 Introduction

Nowadays, router devices are so important with Internet Service Provider (ISP) in core network. Juniper router that is used in SCTV core network enables a wide range of business and residential applications and services (e.g., high-speed transport, Virtual Private Network services, high-speed Internet) [1]. To detect and prevent abnormal operations of Juniper router devices, there are some solutions such as monitoring systems, checking syslog server [2].

Abnormal operations of devices may be cause of routing errors in the network, chassis errors, Distributed Denial-of-service (DDos) attack or some fail processes in router devices. These abnormal operations of router devices are quite reported to syslog server if they are configured. It is possible to detect abnormal operation by inspecting manually these logs. Anomaly detection has been a practical research topic and has great importance in many application domains in university and in industry (e.g., [3], [4], [5]). For conventional small systems, engineers manually define rules or check system logs to detect anomalies based on their domain knowledge. Additionally, they can use regular expression match or keyword searches (e.g., "error", "fail"). However, the extremely large sizes of log data generated (up to million logs) by hundreds of devices make manual analysis impossible.

As a result, automated log analysis methods for anomaly detection of Juniper router devices based on machine learning systems are highly in demand. However, we found no resources about this problem. This paper is the first work about anomaly detection using One-Class SVM for logs of Juniper router devices. We present a new way to get features from log data of Juniper router devices based on importance characteristics of log messages. The standard support vector machine (SVM) [6] is the machine learning method to classify two class or multiple class of data. But the log data of anomaly detection are very special, the abnormal logs are much less than the normal logs. Therefore, the standard SVM does not work well on our task, we use One-Class SVM described in [7] to handle the logs classification.

## 2  Collecting Log Data, Preprocessing and Feature Extraction

Collecting log data for training and testing is necessary because we use the real logs from real router Juniper devices. Logs must be made clean before they can be used for feature extraction.

### 2.1   Collecting Log Data and Text Preprocessing

Juniper router devices routinely generate logs to record device states and runtime information, each including a content indicating what has happened and a timestamp. These devices are configured to send logs to syslog server that always receives data on corresponding port. The log messages have some common characteristics although the format of them is not fixed length. This valuable information could be utilized for anomaly detection and other purposes; thereby logs are collected first and saved as a file in syslog server for further usage. For example, Fig.1 depicts some log messages of Juniper router devices.

```
30-06-2018 07:00:07 AM HCM-Q12-MX5 last message repeated 19 times
30-06-2018 07:00:07 AM HCM-Q12-MX5 inetd[1380]: /usr/sbin/sshd[55491]: exited, status 255
30-06-2018 07:00:10 AM NBH-HED-MX5 tfeb0 MIC(1/0) link 0 SFP syslog throttling: enabling syslogs for receive power alarms and warnings. (0/0)
30-06-2018 07:00:12 AM HCM-Q12-MX5 mib2d[1474]: SNMP_TRAP_LINK_DOWN: ifIndex 533, ifAdminStatus down(1), ifOperStatus down(2), ifName ge-1/1/7
30-06-2018 07:00:15 AM HNI-DDA-MX5 sshd[76567]: error: Received disconnect from 172.16.123.224: 11: disconnected by user
30-06-2018 07:00:15 AM HNI-DDA-MX5 inetd[1368]: /usr/sbin/sshd[76567]: exited, status 255
30-06-2018 07:00:33 AM HNI-BDH-MX5 sshd[15489]: error: Received disconnect from 172.16.123.224: 11: disconnected by user
```

**Fig. 1.** Some log messages of Juniper router devices.

As the example above, a timestamp is the beginning of each message, a name of Juniper router device with the same format and other characteristics: the log messages at syslog server are written in English and comprise digits, lower, upper case letters and many special characters. The raw logs are pre-processed by Python program. Our text normalization procedures are given below:

- Remove timestamps: Remove whole timestamp before each message (both date and time).
- Remove router device's name: Remove device's name.
- Remove digits, special characters: Remove any special character, including punctuations, all digits.
- Replace continuous spaces with a single space: The length of log message depends on how many spaces in log message, so that replacing continuous spaces with a single space is necessary.
- Lower cases: Replace all upper case letters by lower case.

Although timestamp information, which is the time the event happened, is very important according to the SCTV engineers, we still remove it from the log message. After anomaly detection, the abnormal original log messages (including timestamp information) will be sent to the engineers.

### 2.2 Feature Extraction

Some methods used to perform feature extraction for text classification [8-10] are not effective with logs of Juniper router devices. In our works, we use three characteristics of each log message to be three elements of the feature vector: the length of log message, the number of different words and the sum of TF-IDF in log message.

**The Length of Log.**
The length of log message is a significant characteristic. In the process of manually defining abnormal logs, the length of logs shows that the logs with irregular length have a higher probability of being abnormal logs. We denote $S_i$ as the length of log $i$. We use the spaces of each log message to calculate $S_i$.

**The Number of Different Words in Log Message.**
The number of different words in log message: the number of words which are different from dictionary of each log message. Bag-of-Words (BoW) model and the length of log are used to calculate this characteristic. Some of new document representation methods [11-13] are developed based on BoW.

The dictionary is built based on normal logs that we classify from original logs. To get the number of words that are in dictionary, BoW model is the accordant and simple classical model for our purpose. Based on BoW, each vector represents a log message; each element denotes the normalized number of occurrence of a word in log. The words that do not appear in dictionary are not counted by BoW model.

Based on the dictionary and BoW model, each log is converted from text to a vector space. The log data become the matrix that is given by:

$$A = \begin{bmatrix} a_{11} & a_{12} & \cdots & a_{1m} \\ a_{21} & a_{22} & \cdots & a_{2m} \\ \vdots & \vdots & \vdots & \vdots \\ a_{n1} & a_{n2} & \cdots & a_{nm} \end{bmatrix} \quad (1)$$

where $m$ is the length of dictionary, $n$ is the size of log data. Each row of the matrix $A$ represents a log message in log data. The number of words which are different from dictionary of each log message is given by the formula:

$$L_i = S_i - \sum_{j=1}^{m} a_{ij} \quad (2)$$

where $S_i$ is length of the $i$-th log.

**The Sum of TF-IDF.**
Term frequency-inverse document frequency (TF-IDF) is one of the most famous algorithms used in document mining research; it is used for calculating the weight of each word. The word frequency means the number of time a term is repeated in a log message, and Inverse Document Frequency is an algorithm used to calculate the inverse probability of finding a word in log data [14]. Some improvements of feature extraction based on TF-IDF are mentioned in [15, 16].

TF-IDF Formula:

$$g_{ij} = tf_{ij} \times \log\left(\frac{N}{df_j}\right) \quad (3)$$

where $g_{ij}$ is the weight of the word $j$ in the log $i$, $N$ is the total number of log messages, $tf_{ij}$ is the frequency of the word $j$ in the log $i$, $df_j$ is the number of logs containing the word $j$.

The sum of TF-IDF in log is the third element of feature vector. Equation 4 is the summary equation used to calculate the sum of TF-IDF:

$$G_i = \sum_{j=1}^{h} g_{ij} \qquad (4)$$

where $G_i$ is the sum of TF-IDF of log $i$, $h$ is the number of words in log $i$.

The feature vector of log $i$:

$$x_i = (S_i, L_i, G_i) \qquad (5)$$

The feature vectors are the input data of One-Class SVM model, which will be discussed below.

## 3  One-Class SVM

One famous machine learning method is the support vector machine (SVM), which was invented by Vladimir Vapnik and Alexey Ya. Chervonenkis in 1963, and is widely applied for pattern classification and data analyzing. However, Vladimir Vapnik and Corinna Cortes proposed the current standard incarnation in 1993 [6].

The labels associated with training data are based to group anomaly detection techniques for log data into two broad categories: one-class and multi-class anomaly detection technique. The labels in log messages of Juniper router devices are grouped into two types: abnormal log and normal log, so that one-class anomaly detection technique was chosen for our purpose.

For the case of one-class classification, Scholkopf et al. proposed a maximum margin based classifier that is an adaptation of the Support Vector Machine algorithm [7]. The data are separated from the origin by a separating hyperplane $\langle w, z \rangle - \rho = 0$ with maximum margin (where $w$ and $\rho$ are respectively the normal vector of the hyperplane and the distance from the hyperplane to the origin).

The maximum margin from the origin is found by solving the below quadratic optimization problem:

$$\min_{w, \rho, \xi} \quad \frac{1}{2}\|w\|^2 + \frac{1}{vl}\sum_i \xi_i - \rho \qquad (6)$$
$$\text{subject to } (w \cdot \Phi(x_i)) \geq \rho - \xi_i, \ \xi_i \geq 0.$$

where $\xi_i$ are so-called slack variables that are used to model the separation errors. The $v \in (0,1]$ is a parameter that adjusts the balance between maximizing the distance from the origin and the region created by the hyperplane containing most of the data. $\Phi(\cdot)$

is a non-linear projection is evaluated through a kernel function that is used as a mapping from the original feature space to a possibly higher dimensional feature space: $k(x,y) = (\Phi(x) \cdot \Phi(y))$. In our works, we consider the kernel Radial Basis Function (RBF), linear kernel, polynomial kernel and sigmoid kernel. They are expressed respectively by these following equations (7)-(10):

$$k_{(RBF)}(x, y) = e^{-\left(\frac{\|x-y\|^2}{2\sigma^2}\right)} \tag{7}$$

$$k_{(linear)}(x, y) = x^T y + C \tag{8}$$

$$k_{polynomial}(x, y) = \left(\gamma x^T y + C\right)^d \tag{9}$$

$$k_{sigmoid}(x, y) = \tanh\left(\gamma x^T y + C\right) \tag{10}$$

## 4 Experimental Results

In this section, we descript the log datasets, the evaluate method, perform an experimental evaluation and comparison of our anomaly detection method for log data of Juniper router devices based on One-Class SVM model. We have chosen Python programming language for our convenience purpose. Python is a powerful interpreted and popular language and supports some power libraries for data science, machine learning (numpy, matplotlib, scikit-learn, etc) [17- 19].

### 4.1 Log Datasets

Publicly available production logs are scarce data, especially log data of router devices because companies and ISPs rarely publish them to community due to confidential issues. So that we collected log data from real Juniper router devices in core network of SCTV (Saigontourist Cable Television Company). Logs from devices were sent through the internet network to syslog server and saved as txt files. These files can be read directly and easily by most popular programs. The log data contains 12907 log messages and 266 abnormal log messages, which are manually labeled by us and the SCTV experts. Using this log dataset, we evaluated the performance and compared the results of the models. More statistical information of the log dataset is provided in Table 1.

**Table 1.** Summary of log datasets.

| Devices | Data size | Number of messages | Anomalies |
|---|---|---|---|
| Router Juniper | 1,7 Mb | 12 907 | 266 |

After the raw logs are collected using Python script, they are called from our main Python program. Feature extraction is applied to these log messages, each log is represented by a feature vector include: the length of log message, the number of words which are different from dictionary and the sum of TF-IDF value in log message. We choose 60% of log messages as the training data and the remainders as the testing data.

Figure 2 shows the training data and testing data on 3-dimensional feature space. Blue points are the normal data, red points are abnormal data. Due to the fact that the log data come from over a hundred router Juniper devices, we receive a lot of similar log messages. There are some differences between them such as the timestamp, device's name, etc. So a lot of log messages become the same after they are passed the preprocess step that we mentioned above. Because of that, the data points in the figure may look less than reality.

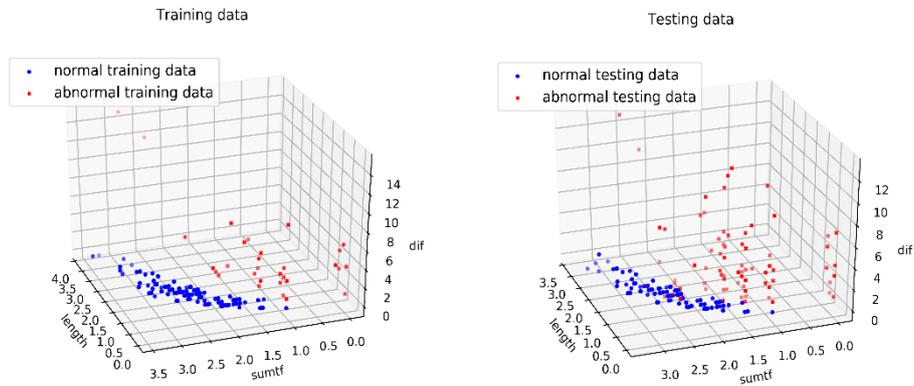

**Fig. 2.** Training data and testing data on 3-dimensional feature space.

Although the visualizations in figure 2 shows that the data are naturally well-separated after they are converted to feature vectors, we still propose the One-Class SVM method to separate the normal log data and abnormal log data automatically, and make sure the model still works efficiently in case there are more abnormal log data appeared when the hardware system is extended.

### 4.2 Method Evaluation

The Precision, Recall and *F*-measure, which are the most commonly used metrics, are used to evaluate the accuracy of One-Class SVM anomaly detection method using different kernels (Radial Basis Function (RBF), linear kernel, polynomial kernel and sigmoid kernel) as we have already the ground truth for the log data. As shown below, Precision shows the percentage of true anomalies among all anomalies detected, Recall measures the percentage of how many real anomalies are detected, and *F*-measure denotes the harmonic mean of precision and recall [20]. They are expressed respectively by the following equations (11) - (13):

$$Precision = \frac{Anomalies\ detected}{Anomalies\ reported} \qquad (11)$$

$$Recall = \frac{Anomalies\ detected}{All\ anomalies} \qquad (12)$$

$$F-measure = \frac{2 \times Precision \times Recall}{Precision + Recall} \qquad (13)$$

### 4.3  Result Evaluation

Based on evaluate methods that we mentioned, we show the results which evaluate of each model applied on training data and testing data. Figure 3 shows the accuracy of anomaly detection on training log data.

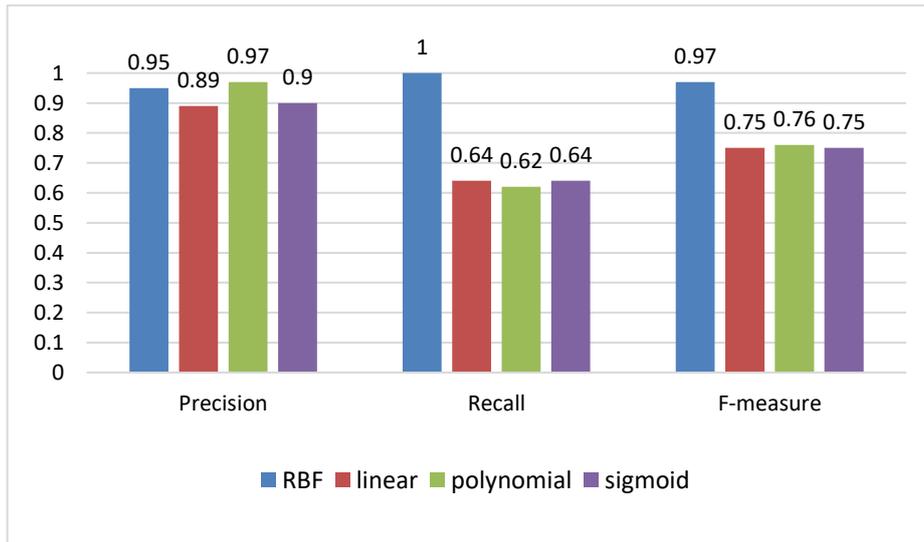

**Fig. 3.** Training accuracy of One-Class SVM method.

Three models (with linear, polynomial, sigmoid kernels) have no good performance on training data with the *F*-measure close to 0.75. We can observe that recall measures of these models are low (close to 0.65). We give priority to minimize the number of abnormal log messages which the model predicted wrongly, One-Class SVM method with RBG kernel is the best model in our case. Recall measure of this model is equal to 1, it shows that the model predicted rightly all abnormal logs. One-Class SVM RBF kernel model's performance on training data is better than others models. However, their accuracy on testing data varies with different kernels.

When applying these models for testing data, three models (with linear, polynomial, sigmoid kernels) become unacceptable. These kernels are not suitable for our purpose.

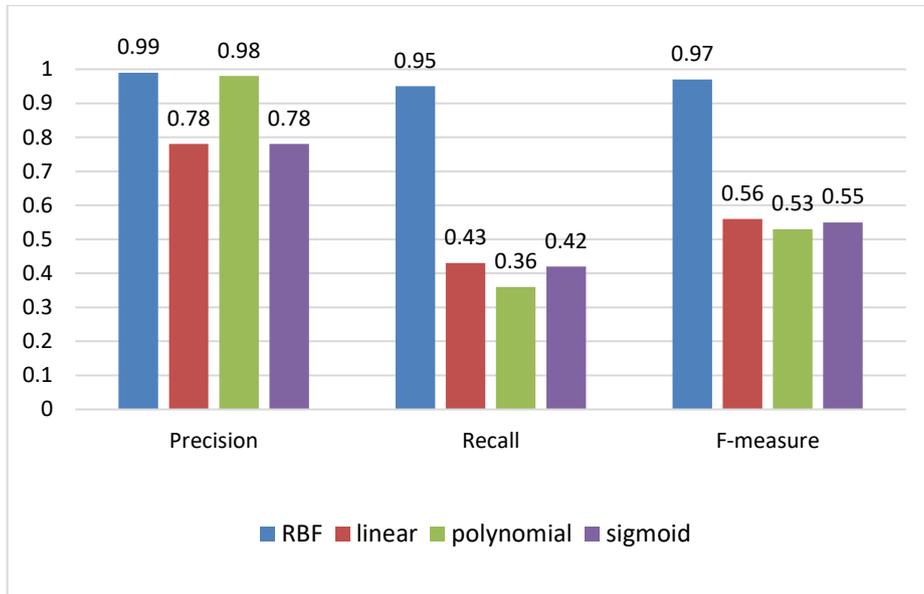

**Fig. 4.** Testing accuracy of One-Class SVM method.

We can observe that One-Class SVM methods with RBF and polynomial kernel achieve high Precision (over 0.95), which implies that normal instances and abnormal instances are well separated by using our feature representation. As we observe on Figure 4, One-Class SVM with RBF kernel has the best results for both training data and testing data.

The decision boundary of One-Class SVM model with RBF kernel is shown on Figure 5.

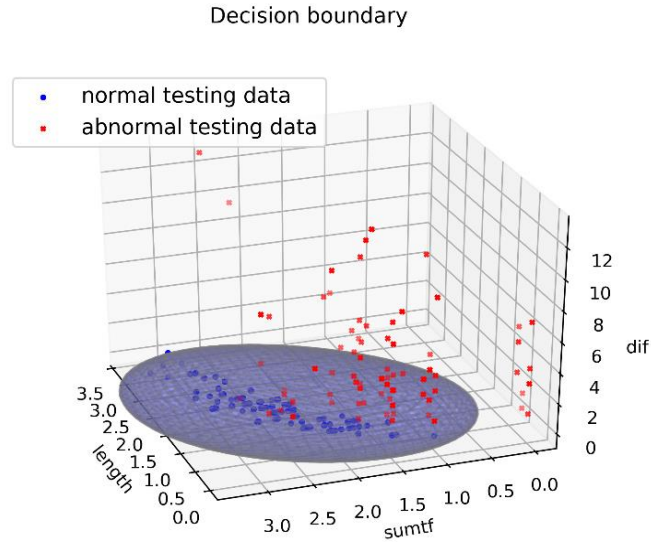

**Fig. 5.** The decision boundary of One-Class SVM model with RBF kernel.

## 5    Conclusion

Logs are widely utilized to detection anomalies in Juniper router device systems. However, traditional anomaly detection that depends heavily on manual log inspection becomes impossible due to the limit of human ability and the sharp increase of log size. To reduce manual effort, automated log analysis and anomaly detection methods have been widely studied in recent years.

In this paper, we successfully created the new feature extraction for log data of Juniper router devices and used the One-Class SVM model with different kernels for anomaly detection. We also compared their accuracy and efficiency on training and testing real log datasets. We find that the One-Class SVM model with RBF kernel has the best accuracy in terms of precision, recall and *F*-measure.

## Acknowledgments

This work was partially supported by the Ministry of Science and Technology, Taiwan, under grant MOST 107-2221-E-006-222. In addition, this work got the encouragement from Posts and Telecommunications Institute of Technology, Vietnam.